\documentclass{article} %
\usepackage{iclr2026_conference,times}

\usepackage{hyperref}
\usepackage{url}
\usepackage{booktabs}
\usepackage{symbols}
\usepackage{amsmath,amssymb}
\usepackage{fancyvrb}
\usepackage{graphicx}
\usepackage{wrapfig}
\usepackage[ruled,linesnumbered]{algorithm2e}
\usepackage[capitalize,noabbrev]{cleveref}
\usepackage{float}
\usepackage{multirow}
\usepackage{adjustbox}

\title{Variational Masked Diffusion Models}

\author{Yichi Zhang\textsuperscript{1},\quad Alex Schwing\textsuperscript{1},\quad Zhizhen Zhao\textsuperscript{1} \\
\\
\textsuperscript{1}University of Illinois Urbana-Champaign\\
}

\iclrfinalcopy %
\begin{document}

\maketitle

\begin{abstract}

Masked diffusion models have recently emerged as a flexible framework for discrete generative modeling. However, a key limitation of standard masked diffusion is its inability to effectively capture dependencies among tokens that are predicted concurrently, leading to degraded generation quality when dependencies among tokens are important. To explicitly model dependencies among tokens, we propose Variational Masked Diffusion (VMD), a framework that introduces latent variables into the masked diffusion process. Through controlled experiments on synthetic datasets, we demonstrate that VMD successfully learns dependencies that conventional masked diffusion fails to capture. We further validate the effectiveness of our approach on Sudoku puzzles and text datasets, where learning of dependencies among tokens improves global consistency. Across these domains, VMD enhances both generation quality and dependency awareness, highlighting the value of integrating variational inference into masked diffusion. Our code is available at: \url{https://riccizz.github.io/VMD}. 

\end{abstract}

\section{Introduction}
\label{sec:intro}
Diffusion-based large language models (DLLMs) represent a significant architectural innovation, emerging as a compelling extension of %
autoregressive models (ARMs). %
This paradigm shift is driven by the inherent limitations of traditional ARMs, which generate tokens sequentially and in a pre-defined order. In contrast, DLLMs offer distinct advantages, including  concurrent token generation, superior output diversity, enhanced global coherence, advanced controllability over the generated text, and presumably easier integration of  heterogeneous data, where a fixed pre-defined sequential token order is harder to justify. 
Recent breakthroughs, exemplified by models like LLaDA~\citep{nie2025largelanguagediffusionmodels}, Mercury~\citep{mercury}, and  Gemini Diffusion~\citep{geminidiffusion}, underscore the increasing viability and promising future  of DLLMs. %

However, their adoption is currently challenged by  %
performance issues in reasoning tasks where tokens are statistically dependent~\citep{li2024promises,xu2025energybased,feng2025theoretical,wu2025fastdllmtrainingfreeaccelerationdiffusion,liu2025plug,song2025ideasinferencetimescalingbenefit,kim2025train}. 
To see this, consider the example  discussed by \citet{song2025ideasinferencetimescalingbenefit} and re-iterated by \citet{wu2025fastdllmtrainingfreeaccelerationdiffusion}: we want to predict the next two words/tokens given the context ``\emph{A poker hand that consists of two English words is:} \underline{\hphantom{a}} \underline{\hphantom{a}}''. Suitable predictions are ``high card,'' ``two pair,'' ``full house,'' or ``straight flush.'' Importantly, a strong dependence exists between these two words. However, concurrent prediction %
in DLLMs does not consider this dependence. This is because prediction in recent DLLMs~\citep{nie2025largelanguagediffusionmodels,yang2025mmadamultimodallargediffusion,you2025lladavlargelanguagediffusion} based on masked diffusion modeling (MDM)~\citep{sahoo2024simpleeffectivemaskeddiffusion,shi2025simplifiedgeneralizedmaskeddiffusion,zheng2025maskeddiffusionmodelssecretly,ou2025absorbingdiscretediffusionsecretly} uses a  %
deep net to compute probability distributions over the vocabulary for each of the masked tokens, and samples from these distributions independently when predicting concurrently. In the poker example, each of the first words ``high,'' ``two,'' ``full,'' and ``straight'' will have roughly a 1/4 probability of occurring. Similarly, the second words ``card,'' ``pair,'' ``house,'' and ``flush'' also have roughly a 1/4 probability of occurring. Due to independent sampling, we are more likely to predict undesirable results than a desirable outcome.

To address this challenge we propose to model a joint token distribution during concurrent prediction by introducing latent variables. A latent variable model enables us to capture any arbitrary joint  distribution across tokens, as opposed to one that factorizes. 
Intuitively, following classic graphical model literature, conditioned on the latent variable we can sample tokens independently, yet, when marginalizing over the latent variable, \ie, when sampling many times, we obtain samples from the proper joint distribution if the model is trained well.
This approach follows the classic variational inference paradigm underlying expectation maximization or variational autoencoders. Hence, we refer to this framework as ``Variational Masked Diffusion'' (VMD).

We  validate the effectiveness of VMD through controlled experiments on synthetic datasets, as well as on Sudoku puzzles and text datasets, where learning of dependencies among tokens improves global consistency. Across these domains, VMD enhances both generation quality and dependency awareness, highlighting the value of integrating variational inference into masked diffusion. %

\section{Preliminaries}
\label{sec:pre}

\begin{figure}[t]
\vspace{-4mm}
    \centering
     \begin{tabular}{cc}
        (a) Training & (b) Sampling \\
    \includegraphics[height=4.6cm]{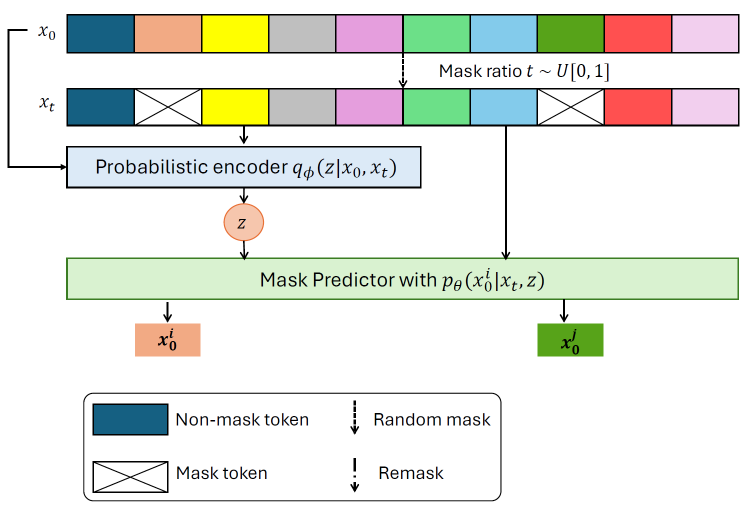} & 
    \includegraphics[height=4.6cm]{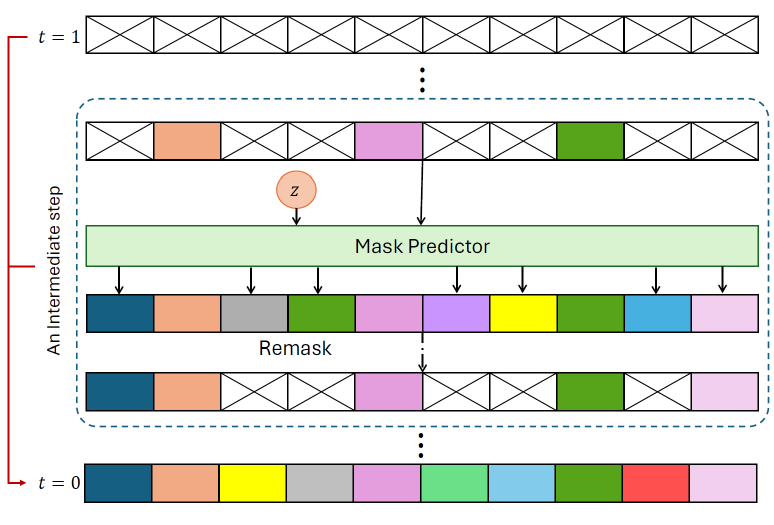} 
    \end{tabular}
    \vspace{-6mm}
    \caption{Conceptual overview of VMD. (a) Training. The encoder and mask predictor are trained on text with random masks applied independently to all tokens at the same ratio $t \sim U[0, 1]$. $x_0^{i}$ and $x_0^{j}$ are conditionally independent given the latent variable $z$ (b) Sampling. Our VMD uses the latent sample $z$ to achieve concurrent mask prediction and recover all tokens at $t = 0$ from the fully masked sequence at $t = 1$ with a flexible remasking strategy. }
    \label{fig:illustration}
    \vspace{-2mm}
\end{figure}

Diffusion models for discrete data were explored recently~\citep{hoogeboom2021argmaxflowsmultinomialdiffusion,austin2023structureddenoisingdiffusionmodels,campbell2022continuoustimeframeworkdiscrete}, and various noise processes were studied. Masked Diffusion Models (MDM)~\citep{sahoo2024simpleeffectivemaskeddiffusion,shi2025simplifiedgeneralizedmaskeddiffusion,zheng2025maskeddiffusionmodelssecretly,ou2025absorbingdiscretediffusionsecretly}, also referred to as absorbing state discrete diffusion models, have gained considerable attention. MDMs use a forward noising process where the original data sequence $x_0 = (x_0^1, \dots, x_0^n)$ consists of $n$ data tokens at diffusion time $t= 0$  which are progressively replaced by a special \texttt{[MASK]} token as diffusion time $t$ increases to $t=1$. Using `$\text{Cat}$' to refer to a categorical distribution, this process is defined by the following transition probability: 
\begin{equation}
q_{t|0}(x_t|x_0) = \prod_{i=1}^n q_{t|0}(x_t^i|x_0^i) = \prod_{i=1}^n \text{Cat}(x_t^i; (1-t)\delta_{x_t^i} + t\delta_{\text{\texttt{[MASK]}}}).
\label{eq:forwardnoising}
\end{equation}

Training of an MDM, as has been shown recently~\citep{sahoo2024simpleeffectivemaskeddiffusion,shi2025simplifiedgeneralizedmaskeddiffusion,zheng2025maskeddiffusionmodelssecretly,ou2025absorbingdiscretediffusionsecretly}, makes use of an objective that can be directly derived from the data likelihood. This leads to the following principled evidence lower bound (ELBO) on the data log likelihood $\log p_\theta(x_0)$:
\begin{equation}
-\log p_\theta(x_0) \leq \int_0^1 \frac{1}{t}\mathbb{E}_{q_{t|0}(x_t|x_0)}\left[\sum_{i:x_t^i = \text{\texttt{[MASK]}}} -\log p_\theta\left(x_0^i|x_t\right)\right]dt.
\label{eq:mdmloss}
\end{equation}
This loss or similar variants are typically used for training a model $p_\theta(x_0^i|x_t)$, which predicts a categorical distribution over the vocabulary for token $x_0^i$ conditioned on partially masked input $x_t$. 

Inference, \ie, the reverse process of the masking defined in \cref{eq:forwardnoising}, is computationally inefficient as it involves modifying only one token per step~\citep{campbell2022continuoustimeframeworkdiscrete,lou2024discretediffusionmodelingestimating}. It is common to apply $\tau$-leaping~\citep{gillespie2001}. This enables us to concurrently unveil multiple masked tokens in a single step from noise level $t$ to noise level $s<t$.

However, as illustrated in \cref{sec:intro} using the poker example, $\tau$-leaping~\citep{gillespie2001} %
in diffusion-based large language models %
faces a trade-off: consider dependence among tokens by decoding one token at a time which is slow, or ignore dependence. This trade-off is suboptimal.

\section{Variational Masked Diffusion (VMD)}
\label{sec:method}
To improve this trade-off, we first discuss in \cref{sec:method:basic} a basic formulation to introduce a latent variable. We then expand this basic variational formulation to block diffusion in \cref{sec:method:block}. We illustrate the benefits of the basic formulation and block diffusion in controlled experiments on synthetic data in \cref{sec:exp:controlled2token,sec:exp:controlled4token}. We study more complex data in \cref{sec:exp:sudoku,sec:exp:text}. 

\begin{figure}[t]
\centering
\begin{minipage}[t]{0.48\textwidth}
\begin{algorithm}[H]
\SetKwComment{Comment}{//}{}
\SetKwInOut{input}{Input}
\SetKwInOut{output}{Output}
\caption{VMD Training}\label{alg:basic:train}
\input{Dataset $\mathcal{D}$, encoder  $\mu_\phi$ and $\sigma_\phi$, decoder  $p_\theta$. \\}
\While{stopping conditions not satisfied}{
 Sample $t \sim {\mathcal{U}[0,1]}$\;
 Sample $x_0 \sim {\mathcal{D}}$\;
 Sample $x_t \sim {q_{t|0}(x_t|x_0)}$\;
 $z = {\mu_\phi(x_t,x_0)+\epsilon\sigma_\phi(x_t,x_0)}$ with $\epsilon\sim\mathcal{N}(0,I)$\;
 Compute $\mathcal{L}_{\text{VMD}}$ in \cref{eq:vmdmobj}\;
 Perform gradient update on $\theta$ and $\phi$\;
}
\output{$\theta$ and $\phi$}
\end{algorithm}

\end{minipage}\hfill
\begin{minipage}[t]{0.48\textwidth}
\begin{algorithm}[H]
\SetKwComment{Comment}{//}{}
\SetKwInOut{input}{Input}
\SetKwInOut{output}{Output}
\caption{VMD Sampling}\label{alg:basic:inf}
\input{Decoder $p_\theta$, fully or partially masked input sequence $x$. }
$z\sim\mathcal{N}(0,I)$, $x_1 = x$\;
\For{$n=N$ {\bfseries to} $1$}{
$t_n \!= \frac{n}{N}, t_{n-1}\!=\frac{n-1}{N}$, and  $x_{t_{n-1}} \!= x_{t_n}$\;\vspace{2.2pt}
 $\hat{x}_0^i = \arg\!\max_v p_\theta(x_0^i = v | x_{t_n}, z), \, 
 \forall i\!:\!x_{t_n}^i\! = \!\text{\texttt{[MASK]}}$\;\vspace{2.2pt}
 $x_{t_{n-1}}^i = \hat{x}_0^i, \forall i\!:\!x_{t_n}^i\! = \!\text{\texttt{[MASK]}}$ and is not remasked\;
}
\output{$x_0$}
\end{algorithm}
\end{minipage}
\vspace{-3mm}
\end{figure}

\subsection{Basic Variational Formulation}
\label{sec:method:basic}
Our basic formulation is contrasted to classic MDM in \cref{fig:illustration}.
Specifically, we introduce a global latent variable. This latent variable permits us to properly characterize arbitrary multi-modal joint probability distributions. For this, we use the categorical mixture model 
\begin{equation}
    p_\theta(x_0^i|x_t) = \int p_\theta(x_0^i|x_t,z)p(z)dz.
    \label{eq:mixture}
\end{equation}
Note that  the latent variable $z$ is global, \ie, it does not depend on the token position $i$. This is important because it enables us to model a joint distribution across tokens. Said differently and following classic graphical model literature: conditioned on the latent variable we can sample tokens $i$ and $j$ independently, yet, when marginalizing over the latent variable %
we obtain samples from the proper joint distribution if the model is trained well. This is formalized as follows:
\begin{equation}
    p_\theta(x_s^i, x_s^j|x_t) = \int p_\theta(x_s^i|x_t, z)\cdot p_\theta(x_s^j|x_t,z) p(z)dz.
\end{equation}

Incorporating the model proposed in \cref{eq:mixture} into the training objective of MDM, given in \cref{eq:mdmloss} yields 
\begin{align}
    &-\log p_\theta(x_0) \leq \mathcal{L}_{\text{VMD}} \nonumber  \\
    & \triangleq \int_0^1 \frac{1}{t}\mathbb{E}_{q_{t|0}(x_t|x_0)}\left[\mathbb{E}_{q_\phi}\left[\sum_{i:x_t^i = \text{\texttt{[MASK]}}} -\log p_\theta(x_0^i|x_t,z)\right] + D_\text{KL}\left(q_\phi(\cdot|x_0,x_t)||p(\cdot)\right)\right]dt.
    \label{eq:vmdmobj}
\end{align}
Here, $p(\cdot)$ is a prior distribution over the latent space, \eg, a standard Gaussian. %
Moreover, $q_\phi(\cdot|x_0,x_t)$ is an approximate posterior parameterized by trainable parameters $\phi$. In a variational autoencoder setting it is often referred to as the encoder while $p_\theta$ is called the decoder. As the approximate posterior is dataset dependent we provide architecture details in \cref{sec:exp} and Appendix~\ref{app:add_exp}. 
The derivation of the negative evidence lower bound (NELBO) is deferred to Appendix~\ref{app:vdm_loss}.  Note that \cref{eq:vmdmobj} maintains the evidence lower bound property. 

For inference, we draw the latent variable $z$ from a standard Gaussian. The trained decoder $p_\theta$ is used to predict all masked tokens at diffusion time $t$ with $\hat x_0^i = \arg\!\max_v p_\theta(x_0^i=v|x_t,z)$ for all $i\!:\!x_t^i\! = \!\text{\texttt{[MASK]}}$. Here, $v$ is a value in the vocabulary. Based on the remasking rules detailed in \cref{sec:remask}, we remask a portion of the tokens and iterate until all  tokens are predicted.   

We summarize the procedure for training and inference in \cref{alg:basic:train} and \cref{alg:basic:inf}, respectively. Training differs from classic masked diffusion model training in minimizing both a cross entropy term and a KL divergence between the approximate posterior and the prior. %
Inference differs from classic masked diffusion model inference in sampling of a single latent variable $z$ for the whole input sequence.

\begin{figure}[t]
\centering
\begin{minipage}[t]{0.45\textwidth}
\begin{algorithm}[H]
\SetKwComment{Comment}{//}{}
\SetKwInOut{input}{Input}
\SetKwInOut{output}{Output}
\caption{Block VMD Training}\label{alg:block:train}
\input{Dataset $\mathcal{D}$, encoder $\mu_\phi$ and $\sigma_\phi$, decoder  $p_\theta$, number of blocks $B$. \\}
\While{stopping conditions not satisfied}{
 Sample $\{t^b\}_{b=1}^B \sim {\mathcal{U}[0,1]}$\;
 Sample $\{x_0^b\}_{b=1}^B \sim {\mathcal{D}}$\;
 Sample $\{x_t^b\}_{b=1}^B \sim {q_{t^b}(x_t^b|x_0^b)}$\;\vspace{1.2pt}
$z^b \!= \!\mu_\phi \left(x_t^b, x_0^{\leq b} \right)+\sigma_\phi\left(x_t^b, x_0^{\leq b}\right)\epsilon$ with $\epsilon\sim\mathcal{N}(0,I), \, \forall b$ \;\vspace{1.2pt}
 Compute $\mathcal{L}_\text{BVMD}$ in \cref{eq:block_nelbo}\;\vspace{1.2pt}
 Perform gradient update on $\theta$ and $\phi$\;
}
\output{$\theta$ and $\phi$}
\end{algorithm}
\end{minipage}\hfill
\begin{minipage}[t]{0.52\textwidth}

\begin{algorithm}[H]
\SetKwComment{Comment}{//}{}
\SetKwInOut{input}{Input}
\SetKwInOut{output}{Output}
\caption{Block VMD Sampling}\label{alg:block:inf}
\input{Decoder $p_\theta$, number of blocks $B$, fully or partially masked input sequence $x$. }
\For{$b=1$ {\bfseries to} $B$}{
 $z^b\sim\mathcal{N}(0,I)$, $x_1^b = x^b$\;
 \For{$n = N$ {\bfseries to} $1$}{
   $t_n \!= \frac{n}{N}, t_{n-1}\!=\frac{n-1}{N}$, and  $x^b_{t_{n-1}} \!= x^b_{t_n}$\;
   $\hat{x}_0^{b, i}\!=\! \arg\!\max_v p_\theta(x_0^i\! =\! v | x_{t_n}^b, x_0^{<b}, z^{\leq b}) \quad \forall i\!:\!x_t^{b, i}\! = \!\text{\texttt{[MASK]}}$\;
    $x_{t_{n-1}}^{b, i} = \hat{x}_0^{b, i}, \forall i\!:\!x_{t_n}^{b, i}\! = \!\text{\texttt{[MASK]}}$ and is not remasked\;
 }
 $x_0 \leftarrow x_0^{1:b-1}\oplus x_0^b$\;
}
\output{$x_0$}
\end{algorithm}
\end{minipage}
\vspace{-5mm}
\end{figure}

\subsection{Block Diffusion Formulation}
\label{sec:method:block}
While a latent variable enables to model dependencies among tokens more accurately, it is challenging to scale this across a large number of tokens. To address this, we follow recent research and combine the strengths of both diffusion and autoregressive models. 

A prime example is the Block Diffusion Language Model (BD3-LM)~\citep{arriola2025block}. This model defines an autoregressive probability distribution over blocks of discrete random variables, with the conditional probability of a block given previous blocks specified by a discrete denoising diffusion model. This hybrid approach effectively overcomes the limitations of both pure diffusion and autoregressive models by supporting flexible-length generation, a historical challenge for DLLMs, and significantly improves inference efficiency by incorporating KV caching and parallel token sampling. 
However, within each block, BD3-LM pursues classic unmasking, \ie, it does not consider the dependencies among tokens that are predicted concurrently. 

To address this, we use a hybrid approach that models blocks of tokens autoregressively and applies the variational diffusion formulation discussed in \cref{sec:method:basic} within each block. 

Formally, the tokens $x$ are grouped in $B$ consecutive blocks, each of length $r$, with total sequence length $L = Br$. For readability, we use  $x^b$ to refer to the block
containing tokens at positions $(b - 1)r+1$ to $br$ with $b \in \{1, \dots, B\}$. We use a latent variable $z^b$ for each block. Using this notation, the data log-likelihood can be factorized as follows:
\begin{equation}
\log p_\theta (x_0 | x_t) = \sum_{b = 1}^B\log p_\theta(x^b_0 | x^b_t, x_0^{<b}) =  \sum_{b = 1}^B\log \int p_\theta(x_0^b |x_t^b, x_0^{<b}, z^{\leq b}) p(z^{\leq b}) dz^{\leq b} , 
\label{eq:mixture_block}
\end{equation}
where $p(z^{\leq b})$ denotes the prior distribution of the latents $z^{\leq b}$. 
The Negative Evidence Lower Bound (NELBO) for the data log-likelihood in \cref{eq:vmdmobj} is changed to 
\begin{align}
\label{eq:block_nelbo}
    -\log p_\theta(x_0) \leq \mathcal{L}_{\text{BVMD}} &\triangleq \sum_{b = 1}^B \int_0^1 \frac{1}{t}\mathbb{E}_{q_{t|0}(x_t|x_0)} \Biggr[\mathbb{E}_{q_\phi}  \left[ \sum_{i:x_t^{b, i} = \text{\texttt{[MASK]}}}-\log p_\theta \left(x_0^{b, i} | x_t^b, x_0^{<b}, z^{\leq b} \right) \right]   \nonumber  \\
    & \quad \quad \quad \quad \quad \quad \quad \quad \quad \quad  + D_\mathrm{KL}\left(q_\phi \left(\cdot | x_t^b, x_0^{\leq b}  \right )\| p \left(\cdot \right ) \right) \Biggr ] dt, 
\end{align}
where $q_\phi \left(\cdot | x_t^b, x_0^{ \leq b} \right)$ is the approximate posterior for the latent parameterized by a neural network with trainable parameters $\phi$. We use  a simple prior $p \left(z^{\leq b} \right) = \mathcal{N}(z^{\leq b}; 0, I)$. The derivation of the bound in~\cref{eq:block_nelbo} is deferred to Appendix~\ref{app:bvdm_loss}.  

Note, when the number of blocks $B = 1$ ($r = L$), 
the model discussed here (\cref{eq:mixture_block}) is identical to the model detailed \cref{eq:mixture}. Moreover, when $B = L$ ($r = 1$), we obtain an autoregressive variational model. Therefore, varying $B$ (or equivalently $r$) allows us to interpolate between an autoregressive variational model and the variational masked diffusion model. 

We summarize training and inference for the variational block diffusion formulation in \cref{alg:block:train} and \cref{alg:block:inf} respectively. During training, we adopt a fully vectorized approach where noisy inputs $x_t$ and clean data $x_0$ are concatenated and processed jointly by the model. An attention mask enforces the dependency structure so that the encoder, when computing the latent $z^b$, only depends on $(x_t^b, x_0^{\leq b})$. %
In contrast, the decoder $p_\theta$ depends on $(x_t^b, x_0^{< b}, z^{\leq b})$.

During inference, we follow block diffusion~\citep{arriola2025block} and apply KV caching for efficient sampling. Each generated block $x_0^b$ is stored in the cache, which means that the prefix context $x_0^{<b}$ used during training corresponds to the accumulated keys and values $(K^{1:b-1},V^{1:b-1})$. The decoder prediction for block $b$, %
therefore, depends only on the current noisy input $x_t^b$, the latents $z^{\leq b}$, and the cached context from earlier generated blocks. In short, training uses ground-truth prefixes while inference reuses cached context, which ensures both correctness and efficiency. 

\subsection{Remasking}
\label{sec:remask}
Early DLLMs~\citep{nie2025largelanguagediffusionmodels} remask the predictions with the 
lowest confidence, \ie,  those with the lowest $p_\theta$ values. %
The remasking strategies considered here include: random remasking and confidence-based  remasking/unmasking~\citep{nie2025largelanguagediffusionmodels, zheng2023reparameterized, kim2025train}. %
The confidence scores used in prior work do not consider the dependence between tokens: they are  token-local metrics. Differently, our variational models provide confidence scores with more global context through the latent variables. Specifically, we consider the following two strategies: %

The first strategy is to check the probability of the selected value at each position~\citep{nie2025largelanguagediffusionmodels, zheng2023reparameterized}. Concretely, if $v_1$  is the most probable value in the vocabulary according to $p_\theta (x^i_0|x_t, z)$, then the confidence for token $i$ is defined as $c^i_{\text{prob.}} = p_\theta(x_0^i = v_1 | x_t, z)$. 

The second strategy uses the top-$K$ probability margin \citep{kim2025train}. The uncertainty of a token is estimated using the absolute difference between the two most probable values at position $i$. If $v_1$ and $v_2$ are the two most probable values in vocabulary according to $p_\theta (x^i_0|x_t, z)$, the confidence of token $i$ is given by
$c^i_{\text{marg.}}\! = \!|p_\theta(x_0^i \!=\! v_1|x_t, z)\! -\! p_\theta(x_0^i\! =\! v_2|x_t, z)|$.

To extend both strategies to block VMD, we replace $p_\theta(x_0^i|x_t, z)$ with $p_\theta(x_0^{b, i} | x_t^b, x_0^{<b}, z^{\leq b})$. 

\section{Experiments}
\label{sec:exp}
We first study efficacy of the proposed Variational Masked Diffusion (VMD) on controlled synthetic data in \cref{sec:exp:controlled2token,sec:exp:controlled4token}. This enables us to carefully assess and visualize the asserted properties of the different formulations. We then present results on Sudoku data in \cref{sec:exp:sudoku}, which has stronger dependencies between tokens than classic text data. Finally, we also present results on text data in \cref{sec:exp:text}, demonstrating that the proposed VMD formulation is on par with prior work.

\subsection{Controlled Synthetic Data with 2 Tokens}
\label{sec:exp:controlled2token}
To examine whether models can capture dependency among tokens, we first evaluate on controlled synthetic datasets where the ground truth distribution is fully specified and easy to visualize. All sequences contain only two tokens, which makes it straightforward to measure whether a model has learned the correct dependency structure. The baseline is a block diffusion model with a small number of DiT blocks. For VMD, the decoder backbone is similar to the baseline to ensure equal inference cost, with an additional lightweight pathway to incorporate latent variables. The encoder uses the same architecture with one quarter of the DiT blocks. A 32-dimensional latent variable is %
injected into the decoder. %
See Appendix~\ref{app:imp:syn} for more implementation details. We report two metrics: (i) accuracy, the proportion of valid sequences among generated samples, and (ii) KL divergence, the distance between the generated distribution and the ground truth distribution.

\begin{table}[t]
\centering
\caption{Results for controlled synthetic data with 2 tokens. We report KL divergence ($\downarrow$) and accuracy ($\uparrow$). While baseline models fail in one-step inference and degenerate to random guessing, VMD successfully captures token dependencies, yielding substantially higher accuracy and lower KL divergence. Best result highlighted in \textbf{bold} font. }
\label{tab:2token1}
\resizebox{1.0\columnwidth}{!}{
\begin{tabular}{llcccc}
\toprule
Experiment & Inference &  MDM (KL $\downarrow$) & MDM (Acc. $\uparrow$) & VMD (KL $\downarrow$) & VMD (Acc. $\uparrow$) \\
\midrule
Deterministic & One-step        & 2.3 & 10.18\% & \textbf{0.082} & \textbf{93.2\%} \\
  & Token-by-token  & \textbf{0.001} & 99.97\% & 0.012 & \textbf{100\%} \\
\midrule
Non-Uniform & One-step        & 2.2 & 11.63\% & \textbf{0.081} & \textbf{93.04\%} \\
  & Token-by-token  & 0.023 & 99.96\% & \textbf{0.009} & \textbf{99.98\%} \\
\bottomrule
\end{tabular}
}

\end{table}

\noindent\textbf{Deterministic dependency.} In the first setting, we consider the following 10 two-token sequences: $\{(k, k+1 \mod 10) \}_{k = 0}^9$ and the data distribution is uniform on the support. Here, the second token is fully determined by the first. From \cref{tab:2token1}, we see that standard masked diffusion fails to capture this dependency when decoding concurrently (one-step inference): when generating both tokens simultaneously, it degenerates to random guessing yielding around 10\% accuracy. In contrast, VMD is able to capture and generate correct pairs reliably. This shows that a latent variable enables the model to represent dependencies among tokens. %

\begin{figure*}[t!]
\vspace{-3mm}
    \centering
    \setlength{\tabcolsep}{0pt}
    {\small
    \begin{tabular}{ccc}
    \includegraphics[height=0.31\linewidth]{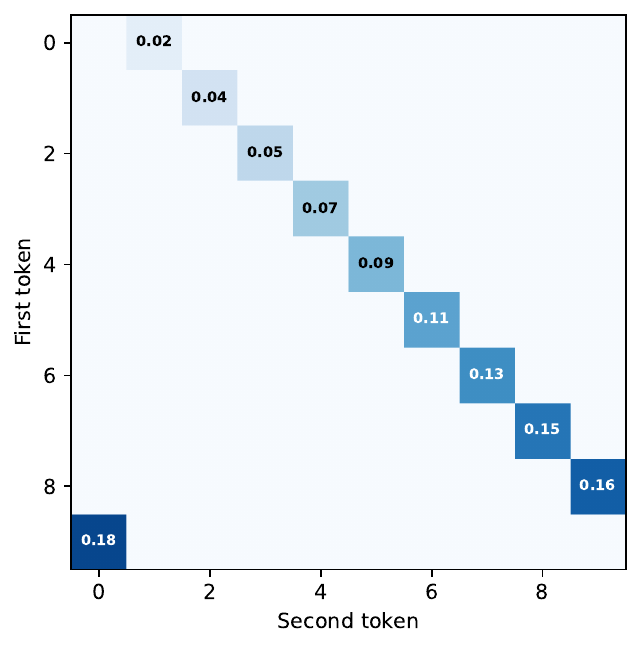}&
    \includegraphics[height=0.31\linewidth]{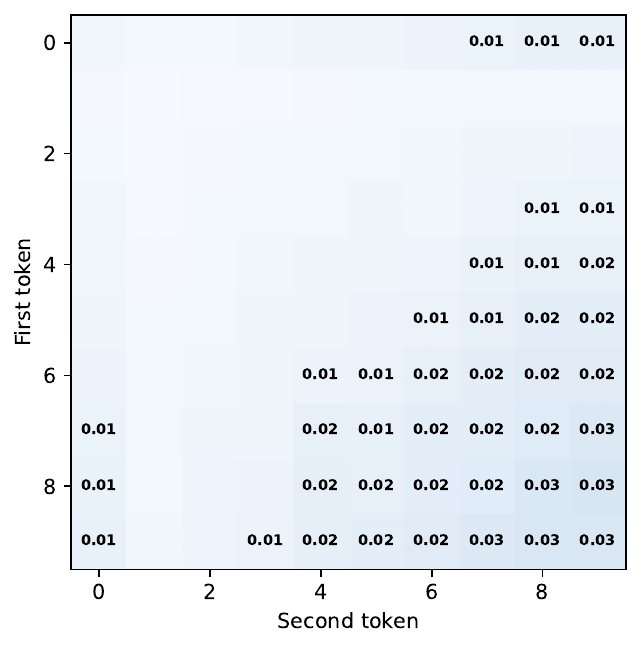}&
    \includegraphics[height=0.314\linewidth]{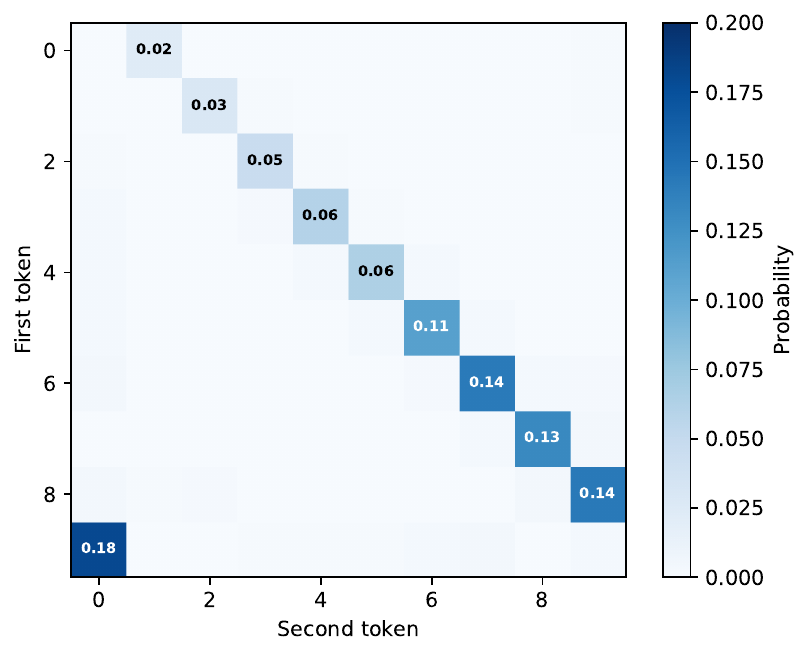} \\[-1mm]
    (a) Ground Truth & (b) Baseline & (c) VMD \\
    \end{tabular}}
    \vspace{-3mm}
    \caption{Results on controlled synthetic data with 2 tokens under the non-uniform setting. (a) Ground truth distribution, (b) baseline masked diffusion with one-step generation, and (c) our VMD with one-step generation. While the baseline degenerates into nearly uniform random guessing and fails to capture the underlying dependency, VMD accurately recovers the true distribution, demonstrating its ability to learn token-to-token correlations beyond conditional independence. }
    \label{fig:non-uniform}
    \vspace{-5mm}
\end{figure*}

\noindent\textbf{Non-uniform distribution.} %
Next, we consider a non-uniform distribution over the support, with $P((k, k+1 \mod 10)) = \frac{k+1}{55}$ for $k\! \in\!\{ 0,...,9\}$. 
While both models achieve near-perfect accuracy with token-by-token generation, KL divergence and heatmap results in \cref{tab:2token1} and \cref{fig:non-uniform} reveal that VMD samples more faithfully from the target distribution. This indicates that VMD not only captures dependencies among tokens but also accurately models the underlying data distribution. 

\begin{wraptable}{r}{0.42\textwidth}
\vspace{-0.7cm}
\caption{KL divergence of experiments on data with varying dependence strength for one-step generation. Higher $p$ values indicate stronger dependence among tokens. Best results shown in \textbf{bold} font. }
\vspace{2mm}
\label{tab:2token2}
\begin{tabular}{lcc}
\toprule
$p$ & MDM (KL $\downarrow$) & VMD (KL $\downarrow$) \\
\midrule
0.3 & 0.155 & \textbf{0.054} \\
0.5 & 0.516 & \textbf{0.062} \\
0.7 & 1.03 & \textbf{0.064} \\
1.0 & 2.31 & \textbf{0.079} \\
\bottomrule
\end{tabular}
\vspace{-20pt}
\end{wraptable}
\noindent\textbf{Varying dependence strength.} Finally, we interpolate between dependent and random pairs. Given a token $x_1$, the second token is set to $(x_1+1) \mod V$ with probability $p$, and to a random alternative with equal probability $\frac{1-p}{V-1}$ otherwise, where $V$ is the vocabulary size. As $p$ decreases from 1 to 0, the token dependency transitions from fully dependent to independent. In this experiment, we cannot use accuracy since every pair of tokens is valid. Therefore, we only report the KL divergence and show heatmaps (Appendix~\ref{app:exp:controlled2token}). Results in \cref{tab:2token2} show that VMD maintains accurate distributional modeling even if the dependence is weak.

Taken together, the three controlled experiments demonstrate: 
(i) for strongly dependent tokens, VMD yields substantial one-step generation improvements; %
(ii) in probabilistic settings, VMD more faithfully models the underlying distribution; 
and (iii) across the full spectrum from strong to weak dependence, VMD consistently models the true data distribution accurately. These results show that VMD successfully learns the token dependency, validating its design. %
Further experimental results can be found in Appendix~\ref{app:exp:controlled2token}.

\subsection{Controlled Synthetic Data with 4 Tokens}
\label{sec:exp:controlled4token}
To further evaluate whether VMD can scale from simple pairwise dependencies to longer sequences, we construct synthetic data with four tokens per sequence. This setting allows us to introduce block structure explicitly and test whether latent variables can simultaneously capture intra-block correlations and preserve cross-block dependencies.   We adopt the same model architecture and evaluation metrics as in the two-token experiments, and set the block size to 2, so that each sequence of length 4 is divided into two blocks. Each block is assigned a 32-dimensional latent variable. %

The first dataset (D1) contains 10 unique sequences: $\{(k, k+1, k+2, k+3 \mod 10) \}_{k = 0}^9$, which have strong token dependence. The second dataset (D2) is  $\{(k, k+1, l, l+1 \mod 10) \}_{k, l = 0}^9$. For D2, the first block of length 2 is independent of the second block of length 2, and the total number of unique sequences is 100. Note that if we independently draw a number for each location, then the total number of unique sequences is $10^4$. Therefore, if we concurrently unmask the digits independently (in one step), the success rate for D1 is $0.1 \%$ and the success rate for D2 is $1.0 \%$ in theory. We provide the results in~\cref{tab:4tokens_main} and observe: when decoding one token at a time, both classic block diffusion as well as our variational formulation achieve almost 100\% prediction accuracy. However, when decoding all four tokens in parallel ($B=1$, NFE~$=1$),  block diffusion behaves like classic MDM and can't do better than random guessing (0.1\% for D1, and 1.1\% for D2), while our approach achieves a much better accuracy of 81.5\% for D1 and 64.4\% for D2. When setting $B\!= \!2$ and NFE~$=2$, we see that the accuracy of block diffusion increases from $0.1\%$ to $10\%$ for D1. The accuracy doesn't change much for D2, since the two consecutive blocks are independent. Our method's accuracy further increases to 91.3\% for D1 and 72.8\% for D2. Also note, for low NFEs, \ie, when generating concurrently, we observe VMD to achieve a lower KL divergence than the baseline. For NFE$~=4$, our generated sequences have larger KL divergence than the baseline, while maintaining a 100\% accuracy. This happens when the generated sequences are valid, but are not as uniformly distributed as the ground truth. 
\begin{table}
\caption{Accuracy ($\uparrow$) (first two rows) and KL divergence ($\downarrow$) (last two rows) for the generated sequences when starting from fully masked tokens. Sequence length $L = 4$. Column `$B\! = \!2$, NFE~$=4$': unmasking one token at a time within each block and autoregressive decoding from block 1 to block 2; Column `$B\! = \!2$, NFE~$=2$': parallel decoding within a block and autoregressive among blocks; Column `$B\! =\! 1$, NFE~$=4$': recovering the sequence token-by-token; Column `$B\! =\! 1$, NFE~$=1$': simultaneous recovery of all tokens. \textbf{Bold} for the best. }
\begin{adjustbox}{width=1.0\textwidth}
    \centering
    \begin{tabular}{cccccccccc}
    \toprule
    & & \multicolumn{4}{c}{D1} & \multicolumn{4}{c}{D2} \\ 
    \cmidrule(lr){3-6} \cmidrule(lr){7-10}
    \multirow{2}{*}{ } & \multirow{2}{*}{ Method} & $B\! =\! 2$  & $B\! = \!2$ & $B \!= \!1$ & $B \!= \!1$ & $B \!= \!2$  & $B \!= \!2$ & $B\! = \!1$ & $B\! =\! 1$ \\
    & & { NFE$= \!4$} & { NFE$=\!2$} & { NFE$=\!4$} & { NFE$=\!1$} & { NFE$=\!4$} & { NFE$=2$} & { NFE$=\!4$} & { NFE$=\!1$} \\
    \midrule
     \multirow{2}{*}{ Acc. ($\uparrow$)} &  {  Block MDM} & {99.4\%} & 10.1\%  & 99.3\% & 0.1 \%  & 99.3\% & 1.1\%  & 99.1 & 1.1\% \\
    & { VMD (ours)} & {\bf 100\%} & {\bf 94.2}\% & {\bf 100}\%  & {\bf 87.3}\% & {\bf 100\%} & {\bf 88.6} \% &  {\bf 100.0}\% & {\bf 80.0} \% \\
    \midrule
     \multirow{2}{*}{  KL ($\downarrow$) } & {  Block MDM} & {\bf 0.007} & 2.298  &  {\bf 0.010} & 9.422 & {\bf 0.018} & 7.656  & {\bf 0.015} & 7.493 \\
    & { VMD (ours)} & 0.012 & {\bf 0.068} & 0.033  & {\bf 0.165} & {\bf 0.018} & {\bf 0.148} & 0.475 & {\bf 0.268} \\
      \bottomrule
    \end{tabular}
    \end{adjustbox}
     \label{tab:4tokens_main}
    \vspace{-3mm}
\end{table}

These results show: VMD extends naturally to block-structured settings. With one latent per block, VMD captures token correlations within blocks, preserves cross-block dependencies when they exist, and avoids inventing dependencies when they do not. Hence, the variational formulation not only addresses the limitations of standard masked diffusion but also integrates seamlessly with block diffusion to provide a principled and flexible framework for modeling dependencies across scales. 

\subsection{Sudoku Data}
\label{sec:exp:sudoku}

Sudoku is a classical logic-based puzzle played on a $9\times9$ grid. The objective is to fill empty cells such that every row, column, and each of the nine $3\times3$ subgrids contains all digits from 1 to 9 exactly once. This formulation naturally imposes global and local dependencies among numbers. The validity of each digit depends not only on its immediate neighbors but also on distant constraints within the same row, column, or subgrid. This makes Sudoku a good benchmark for assessing whether generative models can capture token dependencies. %

\noindent\textbf{Setup.}
We follow the experimental pipeline established by \citet{kim2025train}. Specifically, their work adopts the model and training procedure introduced by \citet{ye2025autoregressiondiscretediffusioncomplex}, but trains on the dataset of \citet{shah2024causallanguagemodelingelicit}. The data consists of 1.8M training puzzles and 0.1M test puzzles. Compared with the smaller 1M-puzzle dataset used by \citet{ye2025autoregressiondiscretediffusioncomplex}, the benchmark of \citet{shah2024causallanguagemodelingelicit} is substantially more challenging and provides a more rigorous test of model generalization. We adopt the same setting and their model as a baseline to ensure comparability and robustness of the evaluation. Following prior work, we represent Sudoku puzzles by flattening each $9\times9$ grid into a sequence of 81 digits. Entries to be filled are denoted by the digit 0. During training, we concatenate the puzzle and its solution, separated by special tokens, yielding an input sequence of length 164. Unlike our other experiments, here we do not employ any block structure. %

\begin{table}[t]
\caption{Sudoku results for different sampling methods and NFE values. VMD consistently improves over the baseline across both sampling schemes. Best results shown in \textbf{bold} font. }
\label{tab:sudoku}
\centering
\begin{tabular}{lcccccc}
\toprule
Model & \multicolumn{3}{c}{Top prob (Accuracy $\uparrow$)} & \multicolumn{3}{c}{Top prob margin (Accuracy $\uparrow$)} \\
\cmidrule(lr){2-4} \cmidrule(lr){5-7}
 & NFE=5 & NFE=10 & NFE=20 & NFE=5 & NFE=10 & NFE=20 \\
\midrule
Baseline & 10.6\% & 14.7\% & 20.4\% & 36.2\% & 78.4\% & 91.1\% \\
VMD     & {\bf 67.7\%} & {\bf 76.4\%} & {\bf 80.9\%} & {\bf 96.9\%} & {\bf 99.0\%} & {\bf 99.7\%} \\
\bottomrule
\end{tabular}
\vspace{-6mm}
\end{table}

\noindent\textbf{Implementation Details for VMD.} The baseline model contains 5.5M parameters. To guarantee comparable inference speed, we keep our decoder backbone approximately the same size as the baseline, while adding a lightweight module to incorporate latent variables. The resulting model has a total of 5.2M parameters. For the encoder, we use the same architecture as the decoder but reduce the number of Transformer layers from 6 to 4. The input sequence is associated with a 128-dimensional latent variable, which is embedded and injected into every DiT block in the decoder via a shared adaptive layernorm module following the design of \citet{chen2025ditairrevisitingefficiencydiffusion}, providing a compact yet expressive representation of dependencies across tokens. 

\noindent\textbf{Evaluation and Results.} We evaluate the performance using the percentage of puzzles that are completely solved (\ie, every cell is correct). This metric directly reflects the ability of a generative model to learn the dependence among tokens, inherent in Sudoku. As shown in \cref{tab:sudoku}, we successfully reproduce the baseline accuracy reported by \citet{kim2025train}. Our VMD model consistently outperforms the baseline using confidence-based remasking with $c_\text{prop.}$ and $c_\text{marg.}$. %
The gains are particularly pronounced at lower NFE, highlighting that VMD can generate valid solutions more efficiently. These results provide strong evidence that VMD learns dependence among tokens that standard masked diffusion fails to capture,  directly supporting our claim.

\subsection{Text Data}
\label{sec:exp:text}

\begin{wraptable}{r}{0.38\textwidth}
\vspace{-10mm}
\caption{Test perplexity (PPL; $\downarrow$) on text8 dataset. \textbf{Bold} for the best. 
}
\label{tab:text}
\centering
\begin{tabular}{ll}
\toprule
&\textbf{PPL ($\downarrow$)}  \\
\midrule
Autoregressive & 2.603 \\
\midrule
SEDD & $\leq$ 3.529 \\
MDLM & $\leq$ 3.498 \\
BD3-LM Block size 4 & $\leq$ 2.873 \\
BD3-LM Block size 8 & $\leq$ 3.126 \\
\midrule
VMD Block size 4 & $\leq$ \textbf{2.858} \\
VMD Block size 8 & $\leq$ 3.125 \\
\bottomrule
\end{tabular}
\vspace{-5mm}
\end{wraptable}
Text data is the most widely used and extensively studied data modality in generative modeling. 
We use the text8 dataset \citep{text8}, which consists of the first 100M characters from Wikipedia. The dataset is preprocessed into a character-level corpus containing 26 English letters and the space symbol, yielding a vocabulary size of 27. Its simplicity and popularity make it a standard benchmark for discrete sequence modeling. 

\noindent\textbf{Implementation Details for VMD.} We segment text sequences into fixed-length chunks of 256 tokens and experiment with two block sizes, 4 and 8, respectively, for block diffusion~\citep{arriola2025block} and our block VMD. For both encoder and decoder, we adapt the architecture of BD3-LM to include latent information. To ensure a fair comparison, the decoder backbone is kept identical to the baseline. For the encoder, similar to the strategy in our Sudoku experiments, we adjust the number of layers so that its architecture mirrors the decoder while maintaining half of its parameter count. A 128-dimensional latent vector is assigned to each block, providing a compact representation of intra-block dependency, while blocks are processed autoregressively using Transformer layers. 

\noindent\textbf{Results.} We compare our method to two state-of-the-art families: (i) strong autoregressive models trained under comparable conditions, and (ii) recent diffusion-based language models, including SEDD~\citep{lou2024discretediffusionmodelingestimating}, MDLM~\citep{sahoo2024simpleeffectivemaskeddiffusion}, and BD3-LM~\citep{arriola2025block}. Test perplexities are reported in \cref{tab:text} as the evaluation metric. Our VMD achieves lower perplexity than standard diffusion baselines (SEDD, MDLM) and provides slight but consistent improvements over BD3-LM at both block sizes. Notably, with block size 4, VMD reaches 2.858 perplexity, which is the best among diffusion-based models. 

These results highlight that text data, with weaker and longer-range dependencies than Sudoku, is a more challenging setting. VMD still achieves consistent gains, indicating that its latent variables are effective at capturing intra-block token dependencies even in large-scale natural language data. VMD provides a principled extension of masked diffusion methods toward competitive text generation, narrowing the gap to autoregressive approaches. 

\section{Related Work}
\label{sec:rel}

\noindent \textbf{Discrete diffusion models. } (Continuous) diffusion models are based on continuous-space Markov chains with
Gaussian transition kernels~\citep{ho2020denoising}. In a similar spirit, discrete diffusion models have emerged
from discrete-space Markov chains~\citep{hoogeboom2021argmaxflowsmultinomialdiffusion}.  \citet{austin2021programsynthesislargelanguage} introduced D3PM, which generalizes the multinomial diffusion model~\citep{hoogeboom2021argmaxflowsmultinomialdiffusion} by going beyond corruption processes with uniform transition probabilities. D3PM formulates a large design space for discrete diffusion models, generalizing and enabling the exploration of new types of corruption processes, including those with absorbing states. The D3PM framework also drew insightful connections between diffusion with absorbing states and masked language models, serving as a critical bridge for the masked diffusion paradigm that would follow. \citet{campbell2022continuous} provided the complete continuous time framework for denoising diffusion
models of discrete data and incorporated the $\tau$-leaping approach for efficient reverse time sampling. Later, \citet{lou2024discretediffusionmodelingestimating}  introduced SEDD, which incorporates a theoretically and practically robust score-entropy objective. Alternatively, several works proposed to perform diffusion over continuous embeddings of discrete tokens~\citep{li2022diffusion,dieleman2022continuous,chen2023analog}. This allows use of continuous diffusion algorithms~\citep{ho2020denoising,song2020denoising}. However, it is challenging to find a good encoding and decoding between  discrete tokens and the continuous latent space~\citep{li2022diffusion}. 

\noindent \textbf{Masked diffusion models. }
Masked diffusion models are a powerful and popular sub-field of discrete diffusion models.
Building upon the absorbing transition kernel, 
\citet{shi2025simplifiedgeneralizedmaskeddiffusion,sahoo2024simpleeffectivemaskeddiffusion} introduce \emph{Masked Diffusion Models} (MDM), which have a simple and principled training recipe, with time dependent mask ratio.  
LLaDA~\citep{nie2025largelanguagediffusionmodels} and LLaDA-V~\citep{you2025lladavlargelanguagediffusion} scaled up the original MDMs and demonstrated their strengths in language modeling. However, all the existing masked diffusion models cannot capture the dependence among concurrently sampled tokens well. Our model addresses this by introducing a latent variable. %

\citet{arriola2025block} introduce block discrete
denoising diffusion language models (BD3-LMs), which interpolate between discrete diffusion
and autoregressive models. It combines the strengths of autoregressive and diffusion models and enables variable-length, high-quality generation while enhancing inference efficiency through KV caching and parallel sampling. Our method further improves  efficiency and accuracy of parallel sampling while using the block autoregressive structure.

\noindent \textbf{Other non-autoregressive models.} 
Efforts to model natural language in a non-autoregressive manner began with BERT~\citep{devlin2019bert}. Such non-causal methods utilize rich text representations. Extending these intuitions further, \citet{hoogeboom2022autoregressive} and \citet{shih2022training} introduced any-order modeling, enabling generation in arbitrary orders. Latent variables have been used in continuous diffusion models, \eg, by~\citet{zhang2025hierarchicalrectifiedflow,zhang2025hierarchicalrectifiedflowmatching,guo2025variationalrectifiedflowmatching}.

\section{Conclusion}
\label{sec:conc}
We address a key limitation of classic masked diffusion modeling: how to consider the dependence among tokens when decoding concurrently. For this, we propose  variational masked diffusion (VMD). It introduces a latent variable into masked diffusion modeling. On data where dependencies among tokens are crucial, VMD significantly improves results upon classic masked diffusion. On data where tokens are largely independent, the proposed formulation yields on par results. All code will be released to ensure reproducibility of the reported results. 

\noindent\textbf{Acknowledgments.} This work is supported in part by NSF grants 2008387, 2045586, 2106825, NIFA award 2020-67021-32799, and OAC 2320345.

\bibliography{ref}
\bibliographystyle{iclr2026_conference}

\clearpage\newpage
\appendix

\section*{Appendix: Variational Masked Diffusion Models}
This appendix is structured as follows: in \cref{app:vdm_loss} we derive the variational masked diffusion (VMD) loss stated in \cref{eq:vmdmobj}; in \cref{app:bvdm_loss} we derive the block diffusion loss stated in \cref{eq:block_nelbo}; in \cref{app:imp} we discuss implementation details; and in \cref{app:add_exp} we provide additional experimental results.

\section{VMD Loss Derivation}
\label{app:vdm_loss}
Here we provide the derivation of the VMD loss stated in \cref{eq:vmdmobj}. For this we use the negative evidence lower bound (NELBO) for  masked diffusion~\citep[Eq.~(10)]{sahoo2024simpleeffectivemaskeddiffusion}, \ie,  
\begin{align}
-\log p_\theta(x_0) \leq  \mathcal{L}^{\infty}_{\text{NELBO}} & \triangleq \int_0^1 \mathbb{E}_{q_{t|0}(x_t|x_0)} \frac{\alpha'_t}{1-\alpha_t}\left[\log p_\theta(x_0|x_t)\right] dt \nonumber \\
& = \int_0^1  \frac{1}{t}\mathbb{E}_{q_{t|0}(x_t|x_0)} \left[-\log p_\theta(x_0|x_t)\right] dt,  \quad \text{with} \quad \alpha_t = 1-t. 
\label{eq:mdm_nelbo}
\end{align}

Next we derive the NELBO for $-\log p_\theta(x_0|x_t)$ as follows:
\begin{align}
-\log p_\theta(x_0|x_t) &= -\log \int p_\theta(x_0|x_t,z)p(z)dz \nonumber \\
 & = -\log \int q_\phi(z | x_0, x_t) \frac{ p_\theta(x_0|x_t,z) p(z)}{ q_\phi(z | x_0, x_t) }dz \nonumber \\
 & \leq  \mathbb{E}_{q_\phi(z | x_0, x_t)} \left [ -\log \frac{ p_\theta(x_0^i|x_t,z)p(z)}{q_\phi(z | x_0, x_t)} \right] \nonumber \\
 & = \mathbb{E}_{q_\phi(z | x_0, x_t)} \left[-\log p_\theta(x_0|x_t,z) \right] - \mathbb{E}_{q_\phi(z | x_0, x_t)} \left [ \log \frac{p(z)}{q_\phi(z | x_0, x_t)} \right] \nonumber \\
 & = \mathbb{E}_{q_\phi(z | x_0, x_t)} \left[-\log p_\theta(x_0|x_t,z) \right] + D_\text{KL}\left(q_\phi(\cdot |x_0,x_t)||p(\cdot)\right).
 \label{eq:pi_nelbo2}
\end{align}

We assume $p_\theta(x_0 | x_t, z) = \prod_{i = 1}^L p_\theta(x_0^i | x_t, z)$. Plugging~\cref{eq:pi_nelbo2} into~\cref{eq:mdm_nelbo}, we obtain
\begin{align}
& -\log p_\theta(x_0)  \leq  \mathcal{L}^{\infty}_{\text{NELBO}}  \nonumber \\
& \leq \int_0^1 \frac{1}{t}\mathbb{E}_{q_{t|0}(x_t|x_0)}\left[\mathbb{E}_{q_\phi}\left[\sum_{i:x_t^i = \text{\texttt{[MASK]}}} -\log p_\theta(x_0^i|x_t,z)\right] + D_\text{KL}\left(q_\phi(\cdot |x_0,x_t)||p(\cdot)\right)\right]dt.  
\end{align}
Here, $p(\cdot)$ is a prior distribution over the latent space, often chosen to be a standard Gaussian, \eg, in a variational autoencoder setting. Moreover, $q_\phi(\cdot|x_0,x_t)$ is an approximate posterior parameterized by trainable parameters $\phi$.

\section{Block VMD Loss Derivation}
\label{app:bvdm_loss}
According to~\cref{eq:mixture_block}, we have
\begin{align}
-\log p_\theta (x_0 | x_t) & =  -\sum_{b = 1}^B\log \int p_\theta(x_0^b |x_t^b, x_0^{<b}, z^{\leq b}) p(z^{\leq b}) dz^{\leq b} , \nonumber \\
& = -\sum_{b = 1}^B \log \int q_\phi(z^{\leq b} | x_t^b, x_0^{\leq b}) \frac{p_\theta(x_0^b |x_t^b, x_0^{<b}, z^{\leq b}) p(z^{\leq b}) }{q_\phi(z^{\leq b} | x_t^b, x_0^{\leq b})} dz^{\leq b}  \nonumber \\
 & \leq -\sum_{b = 1}^B \mathbb{E}_{q_\phi(z^{\leq b} | x_t^b, x_0^{\leq b})} \log \frac{p_\theta(x_0^b |x_t^b, x_0^{<b}, z^{\leq b}) p(z^{\leq b}) }{q_\phi(z^{\leq b} | x_t^b, x_0^{\leq b})} dz^{\leq b}  \nonumber \\
 &= \sum_{b = 1}^B \mathbb{E}_{q_\phi(z^{\leq b} | x_t^b, x_0^{\leq b})} \left[ -\log p_\theta(x_0^b |x_t^b, x_0^{<b}, z^{\leq b}) - \log \frac{p(z^{\leq b}) }{q_\phi(z^{\leq b} | x_t^b, x_0^{\leq b})} \right]  \nonumber \\
 & = \sum_{b = 1}^B \mathbb{E}_{q_\phi(z^{\leq b} | x_t^b, x_0^{\leq b})} \left[ -\log p_\theta(x_0^b |x_t^b, x_0^{<b}, z^{\leq b})\right] + D_{\text{KL}}(q_\phi(\cdot| x^b_t, x_0^{\leq b})\| p(\cdot) ).  \nonumber \\
\label{eq:mixture_block_nelbo}
\end{align}
Here, $p(\cdot)$ is a prior distribution over the latent space, often chosen to be a standard Gaussian. Moreover, $q_\phi(\cdot|x^b_t, x_0^{\leq b})$ is an approximate posterior parameterized by trainable parameters $\phi$. Plugging \cref{eq:mixture_block_nelbo}, into the masked diffusion NELBO given in \cref{eq:mdm_nelbo} yields the bound stated in \cref{eq:block_nelbo}.

\section{Implementation Details}
\label{app:imp}

\subsection{Synthetic Data}
\label{app:imp:syn}

For the synthetic data experiments with both 2-token and 4-token sequences, we use a common model architecture adopted from BD3-LM~\citep{arriola2025block}. The baseline is a block diffusion model with 8 DiT blocks and a hidden size of 64. For the proposed VMD, the decoder backbone remains identical to the baseline with an additional lightweight module to incorporate latent variables to ensure equal inference complexity, while the encoder uses only 2 DiT blocks with all other hyperparameters unchanged. Each block $b$ in the data sequence is associated with a 32-dimensional latent variable $z^b$. The latent is injected into the decoder by fusing it with the positional embedding through a learnable scaling, followed by adaptive layernorm module across all DiT blocks. 

Because the sequence lengths and block sizes are very small, we train with a large batch size of 10,000 for 2,000 steps. We use the Adam optimizer with a fixed learning rate of 1e-3. For the uniform data experiments, we set the KL weight to 4.0 to encourage the latent prior to match a uniform distribution. In all other experiments, when the KL weight is around 1.0, the performance is stable. Parameters are initialized with default PyTorch settings, and no additional regularization is applied beyond the KL term. 

During inference, we adopt the top probability remasking strategy. The number of function evaluations (NFE) depends on the specific experiment and is reported in detail in \cref{sec:exp:controlled2token} and \cref{sec:exp:controlled4token}. All experiments converge within a few minutes on a single NVIDIA L40S GPU. 

\subsection{Sudoku}
\label{app:imp:sudoku}

For the Sudoku experiment, we follow \citet{kim2025train} to ensure a fair comparison by adopting the same dataset, model backbone, and training procedure. Specifically, the baseline uses the HuggingFace \texttt{AutoModelForCausalLM} with GPT-2 configuration ($n_{\text{layer}}=3$, $n_{\text{head}}=12$, embedding size 384). Our model backbone follows the DiT architecture employed by \citet{arriola2025block}. For both the encoder and decoder, we adopt the tiny model configuration provided in their official GitHub repository. We set the encoder and decoder to have 4 and 6 Transformer layers respectively in order to match the model size of the baseline. We further introduce a 128-dimensional latent variable, which is embedded, scaled, and then injected into each of the DiT blocks in the decoder through a shared adaptive layernorm module following the DiT-Air design of \citet{chen2025ditairrevisitingefficiencydiffusion}. 

Data is represented by flattening each Sudoku puzzle into a sequence of 81 tokens. During training, the puzzle and its solution are concatenated into a 164-token sequence (81 puzzle + [SEP] token + 81 solution + [EOS] token), and the model is trained to predict the answer portion conditioned on the puzzle portion. This design implies that the latent variable cannot be obtained by taking a simple mean over the entire sequence dimension. Instead, we compute a learned weighted average over hidden states. First, a gating network produces a score for each position, and positions corresponding to the puzzle are masked out. The remaining scores are normalized with a softmax to produce attention weights, and finally, the latent is obtained as a weighted sum of hidden states according to these weights.  

Training is performed with batch size 1,024, learning rate 1e-3, cosine learning rate scheduler, and 300 epochs. The dataset consists of 1.8M Sudoku puzzles for training and 0.1M for evaluation. During inference, only the puzzle tokens are provided, and the model generates the answer tokens with the remasking strategy described in the main text. 

\subsection{Text}
\label{app:imp:text}
For the text experiments, we adopt a block structure and use the BD3-LM~\citep{arriola2025block} architecture as our baseline. The model processes sequences of length 256 with 8 DiT blocks and 8 attention heads. Our VMD decoder backbone is kept identical to this baseline to ensure equal inference cost, while the encoder mirrors the same architecture but uses only 4 DiT blocks, with all other parameters unchanged. Each block $b$ in the sequence is assigned a 128-dimensional latent variable $z^b$, which is embedded, scaled, and then added to the data embedding in the same manner as in the synthetic and Sudoku experiments.  

Training is performed with batch size 512 using the AdamW optimizer and a learning rate of 3e-4. We employ a constant learning rate schedule with 2,500 warm-up steps.

\section{Additional Experimental Results}
\label{app:add_exp}

\subsection{Controlled Synthetic Data with 2 Tokens}
\label{app:exp:controlled2token}
\begin{figure*}[t]
    \centering
    \setlength{\tabcolsep}{0pt}
    {\small
    \begin{tabular}{ccc}
    \includegraphics[height=0.31\linewidth]{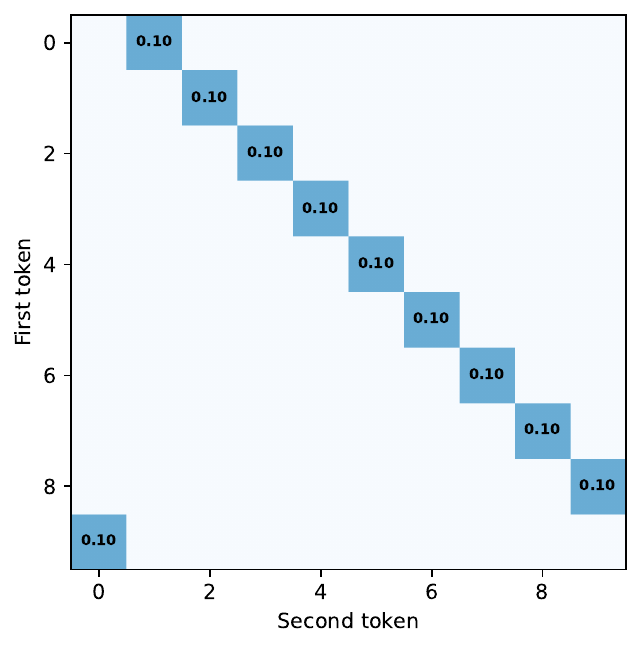}&
    \includegraphics[height=0.31\linewidth]{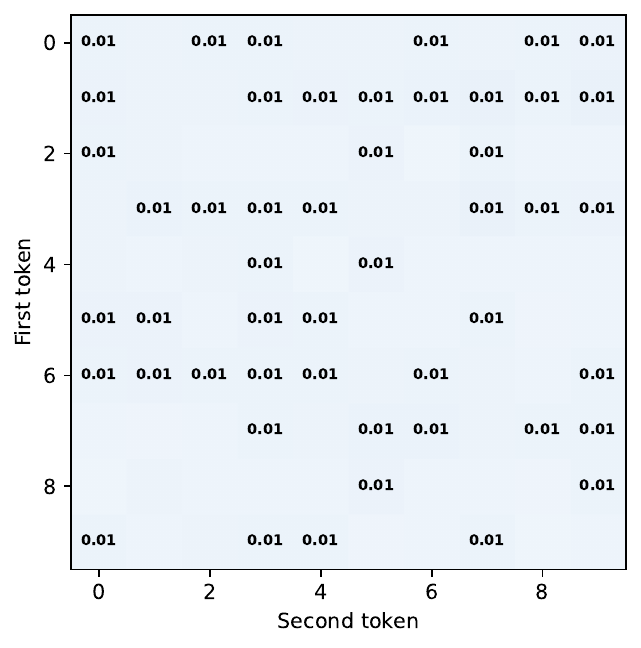}&
    \includegraphics[height=0.314\linewidth]{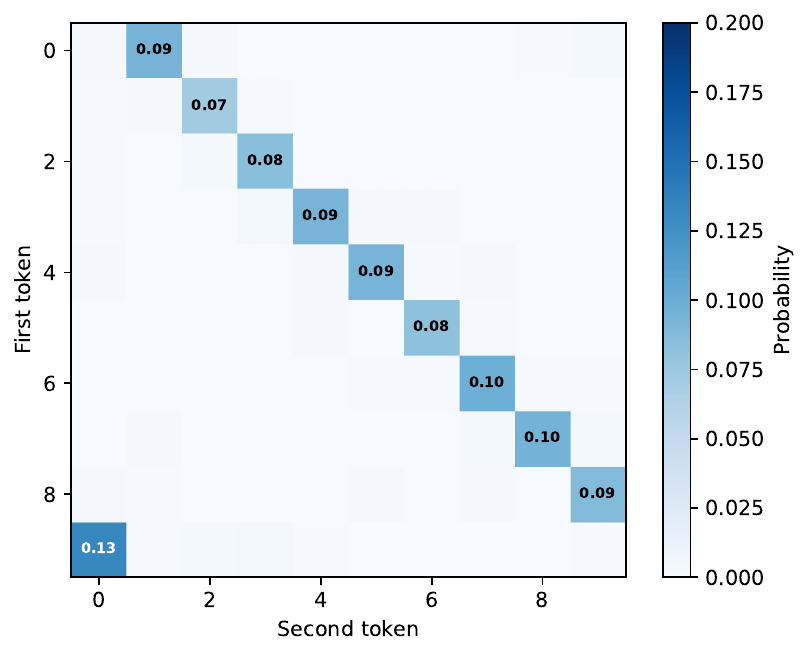} \\
    (a) Ground Truth & (b) MDM & (c) VMD (ours) \\
    \end{tabular}}
    \caption{Results on controlled synthetic data with 2 tokens under the deterministic setting. (a) Ground truth distribution, (b) MDM: baseline masked diffusion with one-step generation, and (c) our VMD with one-step generation. MDM one-step generation is identical to random guessing of each token independently, failing to reflect the true correlations, while VMD closely recovers the ground-truth structure. }
    \label{fig:deterministic}
\end{figure*}

\begin{figure*}[t]
    \centering
    \setlength{\tabcolsep}{0pt}
    {\small
    \begin{tabular}{ccc}
    \includegraphics[height=0.31\linewidth]{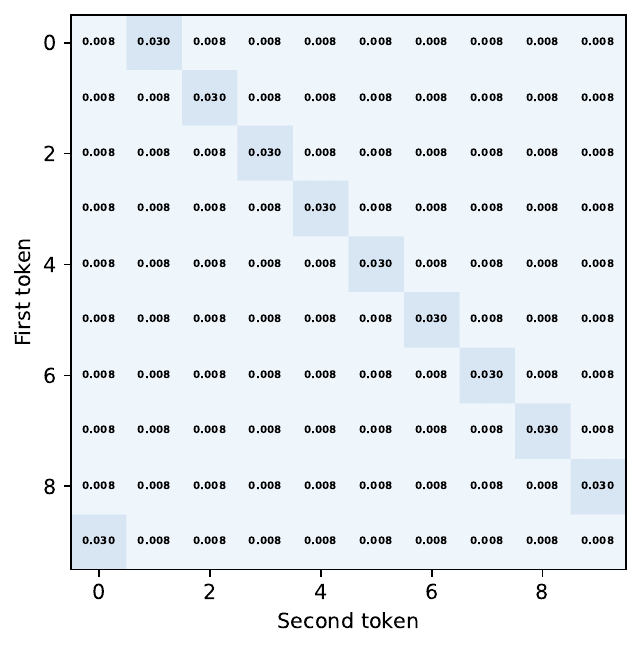}&
    \includegraphics[height=0.31\linewidth]{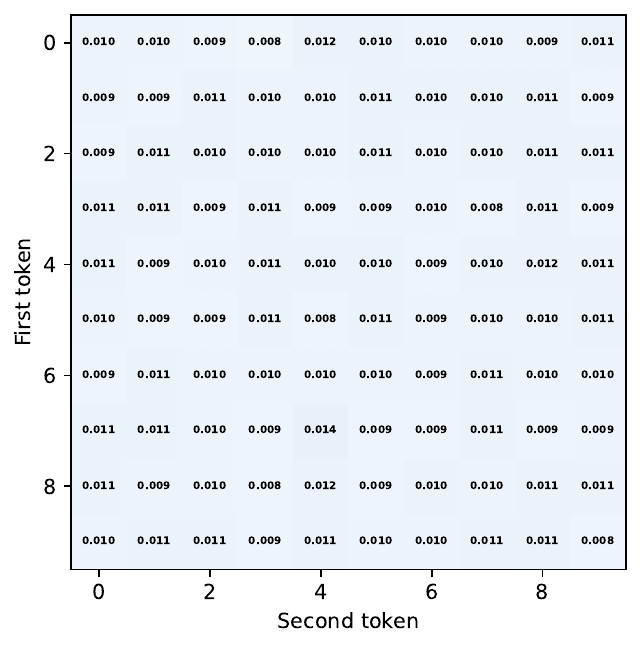}&
    \includegraphics[height=0.314\linewidth]{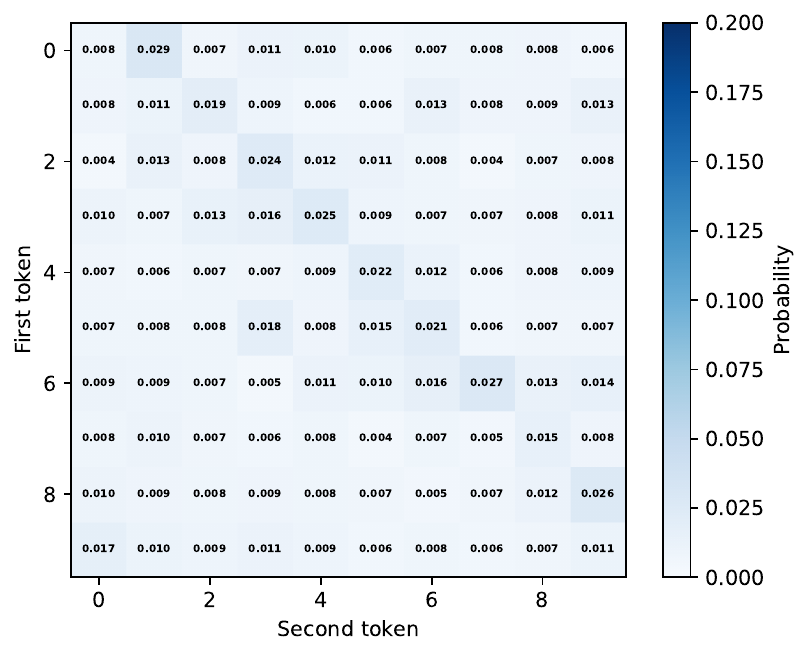} \\
    \includegraphics[height=0.31\linewidth]{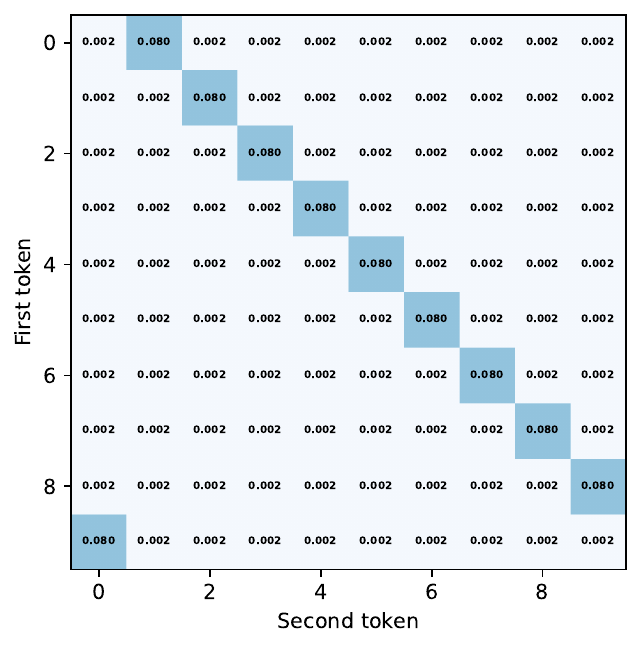}&
    \includegraphics[height=0.31\linewidth]{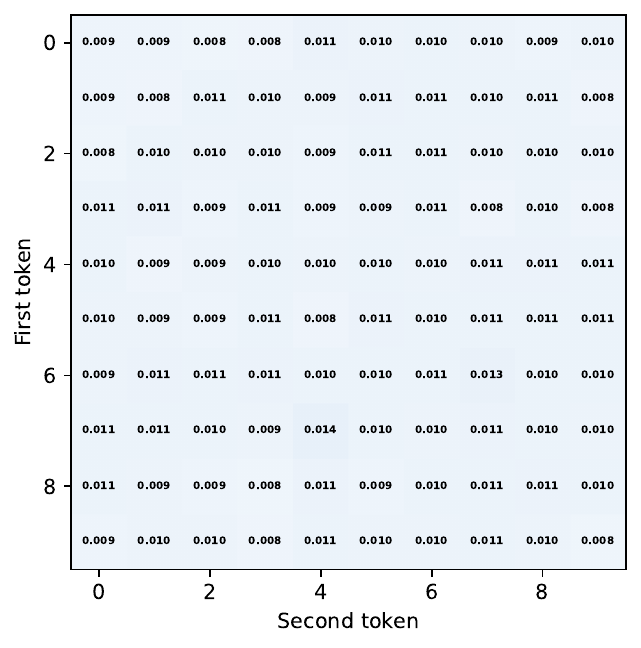}&
    \includegraphics[height=0.314\linewidth]{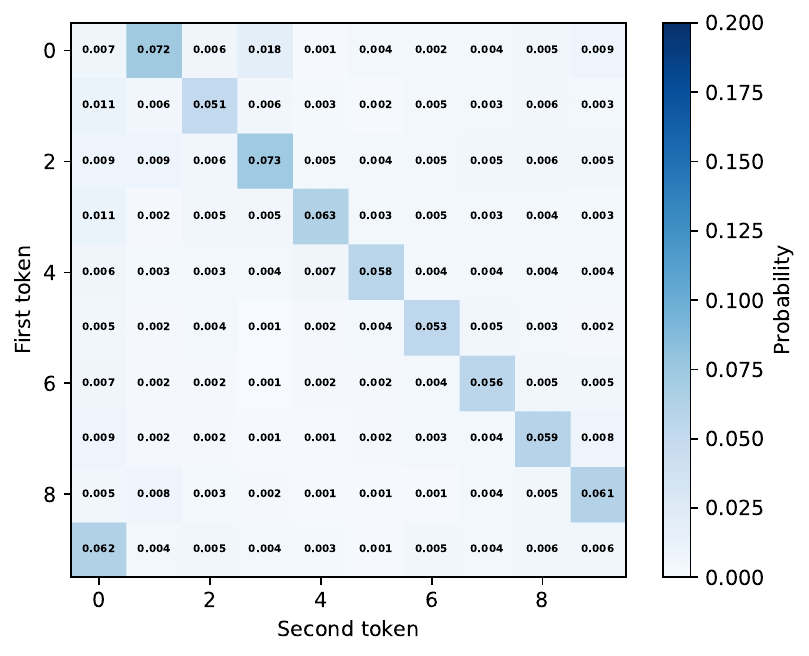} \\
    (a) Ground Truth & (b) MDM & (c) VMD (ours) \\
    \end{tabular}}
    \caption{Results on controlled synthetic data with 2 tokens under the varying correlation setting. (a) Ground truth distribution, (b) MDM: baseline masked diffusion with one-step generation, and (c) our VMD with one-step generation. The first row is for data generated with $p=0.3$, and the second row is for data generated with $p=0.8$. The baseline fails to model the data distribution in one-step generation in both cases, producing nearly uniform predictions, while VMD successfully recovers the underlying distributions. }
    \label{fig:varying-correlation}
\end{figure*}

We present additional experimental results here. \cref{fig:deterministic} and \cref{fig:varying-correlation} present results on controlled synthetic data with 2 tokens under deterministic and varying token dependency settings. In the deterministic setting, the baseline degenerates into nearly uniform predictions and fails to recover the strong token dependence. In contrast, VMD closely matches the ground-truth distribution. In the varying token dependence setting, where a parameter $p$ controls the dependency strength, the baseline again produces noise-like outputs that are insensitive to the true dependency structure. In contrast, VMD adapts to both weak and strong dependence and successfully reproduces the ground-truth distributions. Together, these experiments demonstrate that VMD consistently learns token dependence that standard masked diffusion struggles to capture.

\subsection{Sudoku Data}
\label{app:exp:sudoku}
\begin{figure}[H]
  \centering
  \includegraphics[width=0.7\textwidth]{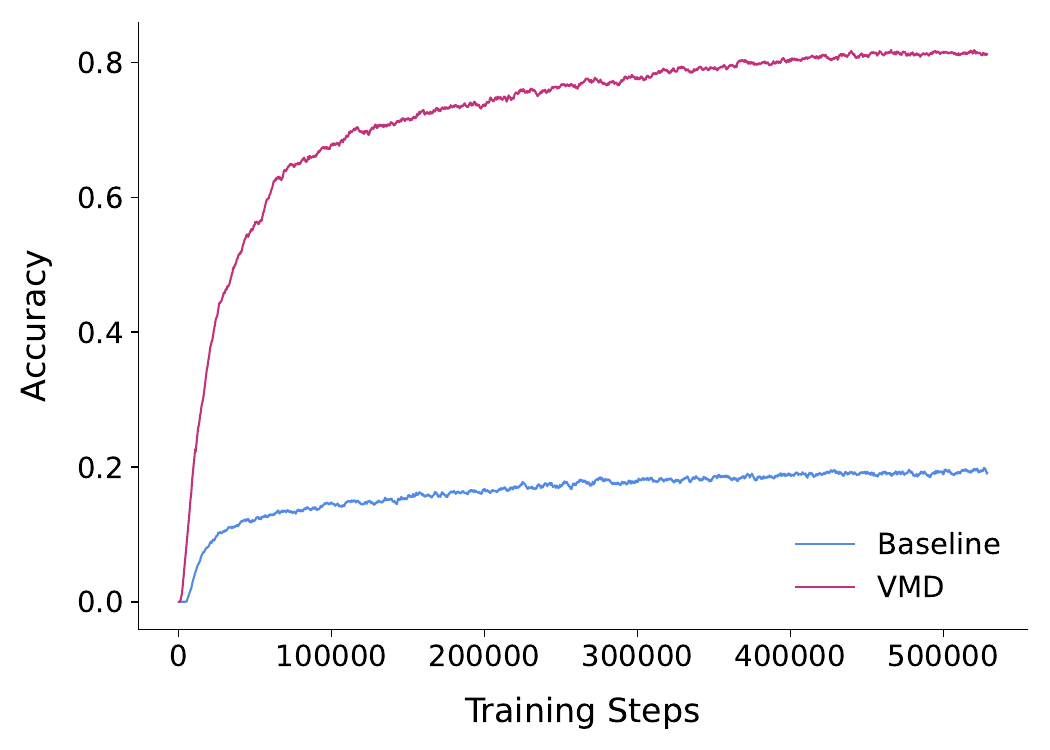}
  \caption{Accuracy ($\uparrow$) during training on the Sudoku puzzle experiment with NFE=20 under the top probability remasking strategy. VMD consistently outperforms the baseline (MDM) across training iterations. The final accuracy reaches 18.97\% for the Baseline and 82.03\% for VMD.}
  \label{fig:sudoku}
\end{figure}

In addition to the final performance reported in the main text, we plot the full training accuracy curves in \cref{fig:sudoku}. The figure shows that VMD consistently outperforms the baseline (MDM) throughout training, not only at convergence but also during earlier stages. This confirms that the gains of VMD are stable across the entire training process rather than arising from a specific checkpoint. We also observe that both models converge at a similar rate, indicating that the improvements of VMD come at no additional optimization cost.

\section{LLM Usage}
While preparing this work, we used a large language model (LLM) to assist with language editing. The LLM's contributions were limited to 
improving the clarity of the text. The core research, experimental design, and all scientific claims remain our original work.

\end{document}